\documentclass[runningheads]{llncs}
\usepackage{graphicx}
\usepackage{tikz}
\usepackage{comment}
\usepackage{amsmath,amssymb} % define this before the line numbering.
\usepackage{color}

\usepackage{enumitem}

\usepackage{adjustbox}
\usepackage{multirow}

\newcommand{\etal}{et~al.}
\newcommand{\tcam}{\mbox{T-CAM}}
\newcommand{\thumos}{\mbox{THUMOS14}}

\newcommand{\actthree}{\mbox{ActivityNet1.3}}
\newcommand{\evideo}{v_i}
\newcommand{\elabel}{\mathbf{y}_i}
\newcommand{\erfeature}{\mathbf{X}_i^r}
\newcommand{\eofeature}{\mathbf{X}_i^o}
\newcommand{\ertcam}{\mathbf{C}_i^r}
\newcommand{\eotcam}{\mathbf{C}_i^o}
\newcommand{\eratt}{\mathbf{A}_i^r}

\newcommand{\ervlfeature}{\mathbf{F}_i^r}

\newcommand{\ercenter}{\mathbf{c}_j^r}

\newcommand{\erwatt}{\mathbf{a}_i^r}

\newcommand{\erwvlfeature}{\mathbf{f}_i^r}

\newcommand{\erafeature}{\mathbf{X}_{i,j}^{ra}}

\newcommand{\eratcam}{\mathbf{C}_i^{ra}}
\newcommand{\eoatcam}{\mathbf{C}_i^{oa}}
\newcommand{\eracenter}{\mathbf{c}_j^{ra}}

\newcommand{\erscore}{\mathbf{s}_i^r}

\newcommand{\efscore}{\mathbf{s}_i^F}
\newcommand{\eftcam}{\mathbf{C}_i^F}
\newcommand{\wrfortcam}{\mathbf{w}^r}
\newcommand{\wofortcam}{\mathbf{w}^o}
\mathchardef\mhyphen="2D
\newcommand\floor[1]{\lfloor#1\rfloor}
\newcommand\ceil[1]{\lceil#1\rceil}

\DeclareMathOperator*{\argmin}{argmin}
\newcommand{\dista}{\mathcal{D}}

\newcommand{\ourtitlenameshort}{Adversarial Background-Aware Loss}
\newcommand{\ourlossnamelong}{Adversarial and Angular Center Loss with a Pair of Triplets}
\newcommand{\ourlossname}{\mbox{A2CL-PT}}
\newcommand{\intlossnamelong}{Angular Center Loss with a Pair of Triplets}
\newcommand{\intlossname}{\mbox{ACL-PT}}
\newcommand{\atclossname}{\mbox{ATCL}}
\newcommand{\tclossname}{\mbox{TCL}}
\newcommand{\clossname}{\mbox{CL}}

\begin{document}
% \renewcommand\thelinenumber{\color[rgb]{0.2,0.5,0.8}\normalfont\sffamily\scriptsize\arabic{linenumber}\color[rgb]{0,0,0}}
% \renewcommand\makeLineNumber {\hss\thelinenumber\ \hspace{6mm} \rlap{\hskip\textwidth\ \hspace{6.5mm}\thelinenumber}}
% \linenumbers
\pagestyle{headings}
\mainmatter
\def\ECCVSubNumber{2121}  % Insert your submission number here

\title{\ourtitlenameshort{} for Weakly-supervised Temporal Activity Localization} % Replace with your title

% INITIAL SUBMISSION 
\begin{comment}
\titlerunning{ECCV-20 submission ID \ECCVSubNumber} 
\authorrunning{ECCV-20 submission ID \ECCVSubNumber} 
\author{Anonymous ECCV submission}
\institute{Paper ID \ECCVSubNumber}
\end{comment}
%******************

% CAMERA READY SUBMISSION
%\begin{comment}
\titlerunning{\ourlossname{} for Weakly-supervised Temporal Activity Localization}
% If the paper title is too long for the running head, you can set
% an abbreviated paper title here
%
\author{Kyle Min \and Jason J. Corso}
%\author{Kyle Min\orcidID{0000-1111-2222-3333} \and Jason J. Corso\orcidID{1111-2222-3333-4444}}
%
\authorrunning{K. Min and J. J. Corso}
% First names are abbreviated in the running head.
% If there are more than two authors, 'et al.' is used.
%
\institute{University of Michigan, Ann Arbor, MI 48109\\
\email{\{kylemin,jjcorso\}@umich.edu}}
%\end{comment}
%******************
\maketitle

\begin{abstract}
Temporally localizing activities within untrimmed videos has been extensively studied in recent years. Despite recent advances, existing methods for weakly-supervised temporal activity localization struggle to recognize when an activity is not occurring. To address this issue, we propose a novel method named \ourlossname{}. Two triplets of the feature space are considered in our approach: one triplet is used to learn discriminative features for each activity class, and the other one is used to distinguish the features where no activity occurs (i.e. background features) from activity-related features for each video. To further improve the performance, we build our network using two parallel branches which operate in an adversarial way: the first branch localizes the most salient activities of a video and the second one finds other supplementary activities from non-localized parts of the video. Extensive experiments performed on \thumos{} and ActivityNet datasets demonstrate that our proposed method is effective. Specifically, the average mAP of IoU thresholds from 0.1 to 0.9 on the \thumos{} dataset is significantly improved from 27.9\% to 30.0\%.

\end{abstract}

\section{Introduction}

\begin{figure}[t]
  \adjustbox{valign=t}{\begin{minipage}[t]{0.42\linewidth}
  \small
    \includegraphics[width=1\linewidth]{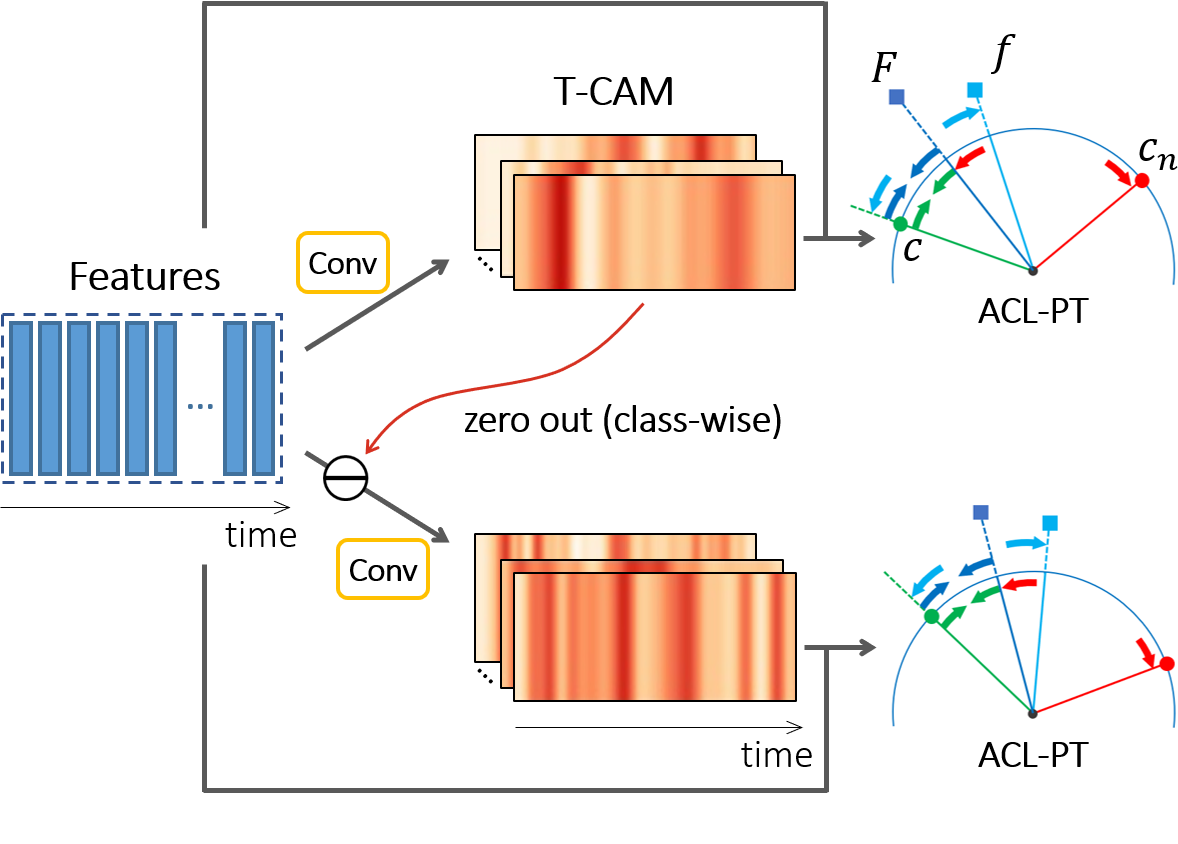}\\[-0.5ex] \hspace*{8.45em}(a)
  \end{minipage}}
  \vline \hspace{0em}
  \adjustbox{valign=t}{\begin{minipage}[t]{0.56\linewidth}
  \small
  \vspace{5ex}
    \includegraphics[width=1\linewidth]{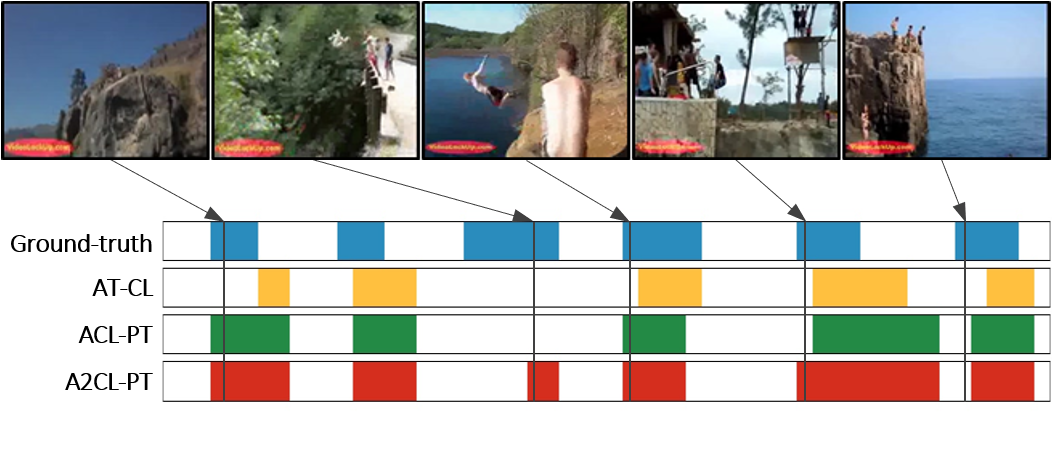}\\[-0.5ex] \hspace*{10.1em}(b)
  \end{minipage}}
  \caption{(a): An illustration of the proposed \ourlossname{}. $F$ and $f$ are aggregated video-level features where $f$ is designed to be more attended to the background features. $c$ is their corresponding center and $c_n$ is the negative center. A triplet of $(F, c, c_n)$ is used to learn discriminative features. We propose to exploit another triplet of $(c, F, f)$ which distinguishes background features from the activity-related features. We call this method of two triplets \intlossname{}. In addition, we design our network with two parallel branches so that the two separate sets of centers can be learned in an adversarial way. We call our final proposed method \ourlossname{}. (b): Sample frames of a video containing \textit{Diving} activity class from \thumos{} dataset~\cite{THUMOS14} and the corresponding results of activity localization. It is shown that our final method \ourlossname{} performs the best.}
  \label{fig1}
\end{figure}

The main goal of temporal activity localization is to find the start and end times of activities from untrimmed videos. Many of the previous approaches are fully supervised: they expect that ground-truth annotations for temporal boundaries of each activity are accessible during training~\cite{shou2016temporal,shou2017cdc,xu2017r,zhao2017temporal,chao2018rethinking,lin2018bsn,long2019gaussian}. However, collecting these frame-level activity annotations is time-consuming and difficult, leading to annotation noise. Hence, a weakly-supervised version has taken foot in the community: here, one assumes that only video-level ground-truth activity labels are available. These video-level activity annotations are much easier to collect and already exist across many datasets \cite{kuehne2011hmdb,soomro2012ucf101,kay2017kinetics,monfortmoments,zhao2019hacs}, thus weakly-supervised methods can be applied to a broader range of situations.

Current work in weakly-supervised temporal activity localization shares a common framework~\cite{liu2019completeness,narayan20193c,nguyen2018weakly,paul2018w,lee2020background}. First, rather than using a raw video, they use a sequence of features extracted by deep networks where the features are much smaller than the raw video in size. Second, they apply a fully-connected layer to embed the pre-extracted features to the task-specific feature space. Third, they project the embedded features to the label space by applying a 1-D convolutional layer to those features. The label space has the same dimension as the number of activities, so the final output becomes a sequence of vectors that represents the classification scores for each activity over time. Each sequence of vectors is typically referred to as CAS (Class Activation Sequence)~\cite{shou2018autoloc} or \tcam{} (Temporal Class Activation Map)~\cite{nguyen2018weakly}. Finally, activities are localized by thresholding this \tcam{}. \tcam{} is sometimes applied with the softmax function to generate class-wise attention. This top-down attention represents the probability mass function for each activity over time.

An important component in weakly-supervised temporal activity localization is the ability to automatically determine background portions of the video where no activity is occurring. For example, \mbox{BaS-Net}~\cite{lee2020background} suggests using an additional suppression objective to suppress the network activations on the background portions. Nguyen \etal{}~\cite{nguyen2019weakly} proposes a similar objective to model the background contents. However, we argue that existing methods are not able to sufficiently distinguish background information from activities of interest for each video even though such an ability is critical to strong temporal activity localization.

To this end, we propose a novel method for the task of weakly-supervised temporal activity localization, which we call \ourlossnamelong{} (\ourlossname{}). It is illustrated in Fig.~\ref{fig1}(a). Our key innovation is that we explicitly enable our model to capture the background region of the video while using an adversarial approach to focus on completeness of the activity learning. Our method is built on two triplets of vectors of the feature space, and one of them is designed to distinguish background portions from the activity-related parts of a video. Our method is inspired by the angular triplet-center loss (\atclossname{})~\cite{li2019angular} originally designed for multi-view 3D shape retrieval. Let us first describe what \atclossname{} is and then how we develop our novel method of \ourlossname{}.

In \atclossname{}~\cite{li2019angular}, a \textit{center} is defined as a parameter vector representing the center of a cluster of feature vectors for each class. During training, the centers are updated by reducing the angular distance between the embedded features and their corresponding class centers. This groups together features that correspond to the same class and distances features from the centers of other class clusters (i.e. negative centers), making the learned feature space more useful for discriminating between classes. It follows that each training sample is a triplet of a feature vector, its center, and a negative center where the feature serves as an anchor.

Inspired by \atclossname{}, we first formulate a loss function to learn discriminative features. \atclossname{} cannot be directly applied to our problem because it assumes that all the features are of the same size, whereas an untrimmed video can have any number of frames. Therefore, we use a different feature representation at the video-level. We aggregate the embedded features by multiplying the top-down attention described above at each time step. The resulting video-level feature representation has the same dimension as the embedded features, so we can build a triplet whose anchor is the video-level feature vector (it is $(F,c,c_n)$ in Fig.~\ref{fig1}(a)). This triplet ensures that the embedded features of the same activity are grouped together and that they have high attention values at time steps when the activity occurs.

More importantly, we argue that it is possible to exploit another triplet. Let us call the features at time steps when some activity occurs \textit{activity features}, and the ones where no activity occurs \textit{background features}. The main idea is that the background features should be distinguished from the activity features for each video. First, we generate a new class-wise attention from \tcam{}. It has higher attention values for the background features when compared to the original top-down attention. If we aggregate the embedded features with this new attention, the resulting video-level feature will be more attended to the background features than the original video-level feature is. In a discriminative feature space, the original video-level feature vector should be closer to its center than the new video-level feature vector is. This property can be achieved by using the triplet of the two different video-level feature vectors and their corresponding center where the center behaves as an anchor (it is $(c, F, f)$ in Fig.~\ref{fig1}(a)). The proposed triplet is novel and will be shown to be effective. Since we make use of a pair of triplets on the same feature space, we call it \intlossnamelong{} (\intlossname{}).

To further improve the localization performance, we design our network to have two parallel branches which find activities in an adversarial way, also illustrated in Fig.~\ref{fig1}(a). Using a network with a single branch may be dominated by salient activity features that are too short to localize all the activities in time. We zero out the most salient activity features localized by the first branch for each activity so that the second (adversarial) branch can find other supplementary activities from the remaining parts of the video. Here, each branch has its own set of centers which group together the features for each activity and one 1-D convolutional layer that produces \tcam{}. The two adversary \tcam{}s are weighted to produce the final \tcam{} that is used to localize activities. We want to note that our network produces the final \tcam{} with a single forward pass so it is trained in an end-to-end manner. We call our final proposed method \ourlossnamelong{} (\ourlossname{}). It is shown in Fig.~\ref{fig1}(b) that our final method performs the best.

There are three main contributions in this paper:

\begin{itemize}[noitemsep,nolistsep]
   \item We propose a novel method using a pair of triplets. One facilitates learning discriminative features. The other one ensures that the background features are distinguishable from the activity-related features for each video.
   \item We build an end-to-end two-branch network by adopting an adversarial approach to localize more complete activities. Each branch comes with its own set of centers so that embedded features of the same activity can be grouped together in an adversarial way by the two branches.
   \item We perform extensive experiments on \thumos{} and ActivityNet datasets and demonstrate that our method outperforms all the previous state-of-the-art approaches.
\end{itemize}

\section{Related Work}

Center loss (\clossname{})~\cite{wen2016discriminative} is recently proposed to reduce the intra-class variations of feature representations. \clossname{} learns a center for each class and penalizes the Euclidean distance between the features and their corresponding centers. Triplet-center loss (\tclossname{})~\cite{he2018triplet} shows that using a triplet of each feature vector, its corresponding center, and a nearest negative center is effective in increasing the inter-class separability. \tclossname{} enforces that each feature vector is closer to its corresponding center than to the nearest negative center by a pre-defined margin. Angular triplet-center loss (\atclossname{})~\cite{li2019angular} further improves \tclossname{} by using the angular distance. In \atclossname{}, it is much easier to design a better margin because it has a clear geometric interpretation and is limited from 0 to $\pi$.

\mbox{BaS-Net}~\cite{lee2020background} and Nguyen \etal{}~\cite{nguyen2019weakly} are the leading state-of-the-art methods for weakly-supervised temporal activity localization. They take similar approaches to utilize the background portions of a video. There are other recent works without explicit usage of background information. Liu \etal{}~\cite{liu2019completeness} utilizes multi-branch network where \tcam{}s of these branches differ from each other. This property is enforced by the diversity loss: the sum of the simple cosine distances between every pair of the \tcam{}s. 3C-Net applies an idea of \clossname{}, but the performance is limited because \clossname{} does not consider the inter-class separability.% 3C-Net improves the performance by using additional training annotations for activity count (the number of occurrences of each activity per video).

Using an end-to-end two-branch network that operates in an adversarial way is proposed in Adversarial Complementary Learning (ACoL)~\cite{zhang2018adversarial} for the task of weakly-supervised object localization. In ACoL, object localization maps from the first branch are used to erase the salient regions of the input feature maps for the second branch. The second branch then tries to find other complementary object areas from the remaining regions. To the best of our knowledge, we are the first to merge the idea of ACoL with center loss and to apply it to weakly-supervised temporal activity localization.

\section{Method}

\begin{figure}[t]
  \centering
  \includegraphics[width=\linewidth]{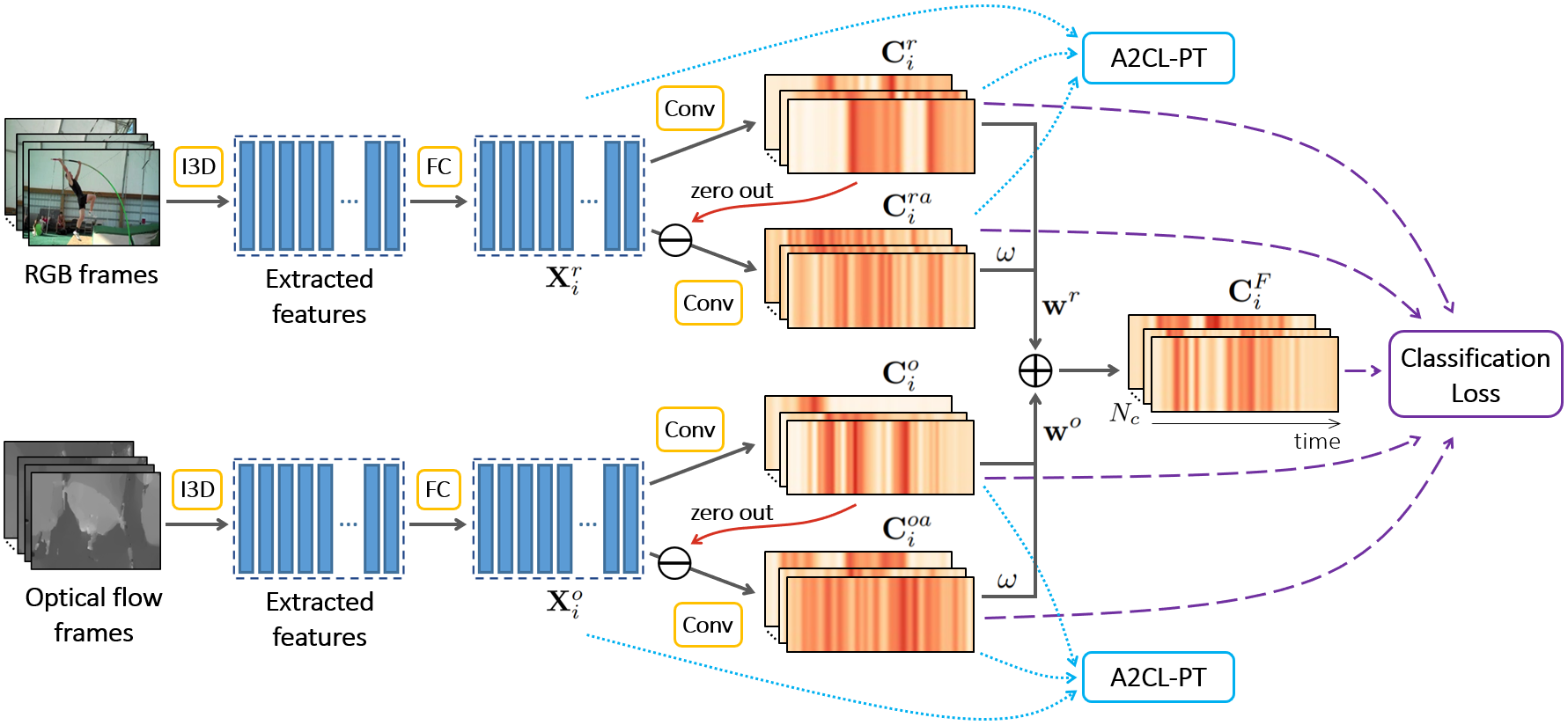}
  \caption{An illustration of our overall architecture. It consists of two streams (RGB and optical flow), and each stream consists of two (first and adversarial) branches. Sequences of features are extracted from two input streams using pre-trained I3D networks~\cite{carreira2017quo}. We use two fully-connected layers with ReLU activation (FC) to compute the embedded features $\erfeature, \eofeature$. Next, \tcam{}s $\ertcam, \eotcam$ are computed by applying 1-D convolutional layers (Conv). The most salient activity features localized by the first branch are zeroed out for each activity class, and the resulting features are applied with different 1-D convolutional layers (Conv) to produce $\eratcam, \eoatcam$. Using the embedded features $\erfeature, \eofeature$ and \tcam{}s $\ertcam, \eotcam, \eratcam, \eoatcam$, we compute the term of \ourlossname{} (Eq.~\ref{eq11}). The final \tcam{} $\eftcam$ is computed from the four \tcam{}s and these \tcam{}s are used to compute the loss function for classification (Eq.~\ref{eq14}).}
  \label{fig2}
\end{figure}

The overview of our proposed method is illustrated in Fig.~\ref{fig2}. The total loss function is represented as follows:
\begin{equation}
\label{eq1}
\mathcal{L} = \alpha\mathcal{L}_{\text{\ourlossname{}}}+\mathcal{L}_{\text{CLS}}
\end{equation}
where $\mathcal{L}_{\text{\ourlossname{}}}$ and $\mathcal{L}_{\text{CLS}}$ denote our proposed loss term and the classification loss, respectively. $\alpha$ is a hyperparameter to control the weight of \ourlossname{} term.
In this section, we describe each component of our method in detail.

\subsection{Feature Embedding}
Let us say that we have $N$ training videos ${\{\evideo\}_{i=1}^N}$. Each video $\evideo$ has its ground-truth annotation for video-level label ${\elabel \in \mathbb{R}^{N_c}}$ where $N_c$ is the number of activity classes. ${\elabel(j)=1}$ if the activity class $j$ is present in the video and ${\elabel(j)=0}$ otherwise. We follow previous works~\cite{paul2018w,narayan20193c} to extract the features for both RGB and optical flow streams. First, we divide $\evideo$ into non-overlapping 16-frame segments. We then apply I3D~\cite{carreira2017quo} pretrained on Kinetics dataset~\cite{kay2017kinetics} to the segments. The intermediate $D$-dimensional ($D=1024$) outputs after the global pooling layer are the pre-extracted features. For the task-specific feature embedding, we use two fully-connected layers with ReLU activation. As a result, sequences of the embedded features ${\erfeature,\eofeature \in \mathbb{R}^{D \times l_i}}$ are computed for RGB and optical flow stream where $l_i$ denotes the temporal length of the features of the video $\evideo$.

\subsection{\intlossnamelong{} (\intlossname{})}
\label{subsec:adot}
For simplicity, we first look at the RGB stream. The embedded features $\erfeature$ are applied with a 1-D convolutional layer. The output is \tcam{} ${\ertcam \in \mathbb{R}^{N_c \times l_i}}$ which represents the classification scores of each activity class over time. We compute class-wise attention ${\eratt \in \mathbb{R}^{N_c \times l_i}}$ by applying the softmax function to \tcam{}:
\begin{equation}
\label{eq2}
\eratt(j,t)=\frac{\exp\big(\ertcam(j,t)\big)}{\textstyle \sum_{t'=1}^{l_i}\exp\big(\ertcam(j,t')\big)}
\end{equation}
where ${j \in \{1,...,N_c\}}$ denotes each activity class and $t$ is for each time step. Since this top-down attention represents the probability mass function of each activity over time, we can use it to aggregate the embedded features $\erfeature$:
\begin{equation}
\label{eq3}
\ervlfeature(j)=\sum_{t=1}^{l_i}\eratt(j,t)\erfeature(t)
\end{equation}
where $\ervlfeature(j) \in \mathbb{R}^D$ denotes a video-level feature representation for the activity class $j$. Now, we can formulate a loss function that is inspired by \atclossname{}~\cite{li2019angular} on the video-level feature representations as follows:
\begin{equation}
\label{eq4}
%\mathcal{L}_{\text{\atclossname{}}}^r = \frac{1}{N}\sum_{i=1}^N \sum_{j:\elabel(j)=1} \max \Big(0, D\big(\ervlfeature(j), \ercenter\big)-\min\limits_{k\neq j}D\big(\ervlfeature(j), \mathbf{c}_k^r\big) + m_1 \Big)
\mathcal{L}_{\text{\atclossname{}}}^r = \frac{1}{N}\sum_{i=1}^N \sum_{j:\elabel(j)=1} \max \Big(0, \dista \big(\ervlfeature(j), \ercenter\big)-\dista \big(\ervlfeature(j), \mathbf{c}_{n_{i,j}^r}^r\big) + m_1 \Big)
\end{equation}
where $\ercenter \in \mathbb{R}^D$ is the center of activity class $j$, $n_{i,j}^r=\argmin\limits_{k\neq j} \dista \big( \ervlfeature(j), \mathbf{c}_k^r\big)$ is an index for the nearest negative center, and $m_1 \in [0, \pi]$ is an angular margin. It is based on the triplet of $(\ervlfeature(j),\ercenter,\mathbf{c}_{n_{i,j}^r}^r)$ that is illustrated in Fig.~\ref{fig1}(a). Here, $\dista(\cdot)$ represents the angular distance:
\begin{equation}
\label{eq5}
\dista \big(\ervlfeature(j), \ercenter\big)=\arccos \bigg( \frac{\ervlfeature(j) \cdot \ercenter}{\| \ervlfeature(j) \|_2 \| \ercenter \|_2} \bigg)
\end{equation}
Optimizing the loss function of Eq.~\ref{eq4} ensures that the video-level features of the same activity class are grouped together and that the inter-class variations of those features are maximized at the same time. As a result, the embedded features are learned to be discriminative and \tcam{} will have higher values for the activity-related features.

For the next step, we exploit another triplet. We first compute a new class-wise attention ${\erwatt \in \mathbb{R}^{N_c \times l_i}}$ from \tcam{}:
\begin{equation}
\label{eq6}
\erwatt(j,t)=\frac{\exp\big(\beta \ertcam(j,t)\big)}{\textstyle \sum_{t'=1}^{l_i}\exp\big(\beta\ertcam(j,t')\big)}
\end{equation}
where $\beta$ is a scalar between 0 and 1. This new attention still represents the probability mass function of each activity over time, but it is supposed to have lower values for the activity features and higher values for the background features when compared to the original attention $\eratt$. Therefore, if we aggregate the embedded features $\erfeature$ using $\erwatt$, the resulting new video-level feature $\erwvlfeature$ should attend more strongly to the background features than $\ervlfeature$ is. This property can be enforced by introducing a different loss function based on the new triplet of $(\ercenter, \ervlfeature(j), \erwvlfeature(j))$ that is also illustrated in Fig.~\ref{fig1}(a):
\begin{equation}
\label{eq7}
\mathcal{L}_{\text{NT}}^r = \frac{1}{N}\sum_{i=1}^N \sum_{j:\elabel(j)=1} \max \Big(0, \dista \big(\ervlfeature(j), \ercenter\big)-\dista \big(\erwvlfeature(j), \ercenter\big) + m_2 \Big)
\end{equation}
where the subscript NT refers to the new triplet and $m_2 \in [0, \pi]$ is an angular margin. Optimizing this loss function makes the background features more distinguishable from the activity features. Merging the two loss functions of Eq.~\ref{eq4} and Eq.~\ref{eq7} gives us a new loss based on a pair of triplets, which we call \intlossnamelong{} (\intlossname{}):
\begin{equation}
\label{eq8}
\mathcal{L}_{\text{\intlossname{}}}^r = \mathcal{L}_{\text{\atclossname{}}}^r+\gamma\mathcal{L}_{\text{NT}}^r
\end{equation}
where $\gamma$ is a hyperparameter denoting the relative importance of the two losses.

Previous works on center loss~\cite{wen2016discriminative,he2018triplet,li2019angular} suggest using an averaged gradient (typically denoted as $\Delta \ercenter$) to update the centers for better stability. Following this convention, the derivatives of each term of Eq.~\ref{eq8} with respect to the centers are averaged. For simplicity, we assume that the centers have unit length. Refer to the supplementary material for general case without such assumption. Let $\tilde{\mathcal{L}}_{\text{\atclossname{}}_{i,j}}^r$ and $\tilde{\mathcal{L}}_{\text{NT}_{i,j}}^r$ be the loss terms inside the max operation of the $i$-th sample and of the $j$-th activity class as follows:
\begin{equation}
\label{eqc1}
\tilde{\mathcal{L}}_{\text{\atclossname{}}_{i,j}}^r = \dista \big(\ervlfeature(j), \ercenter\big)-\dista \big(\ervlfeature(j), \mathbf{c}_{n_{i,j}^r}^r\big) + m_1
\end{equation}
\begin{equation}
\label{eqc2}
\tilde{\mathcal{L}}_{\text{NT}_{i,j}}^r = \dista \big(\ervlfeature(j), \ercenter\big)-\dista \big(\erwvlfeature(j), \ercenter\big) + m_2
\end{equation}
Next, let $\mathbf{g}_{1_{i,j}}^r$ and $\mathbf{g}_{2_{i,j}}^r$ be the derivatives of Eq.~\ref{eqc1} with respect to $\ercenter$ and $\mathbf{c}_{n_{i,j}^r}^r$, respectively; and let $\mathbf{h}_{i,j}^r$ be the derivative of Eq.~\ref{eqc2} with respect to $\ercenter$. For example, $\mathbf{g}_{1_{i,j}}^r$ is given by:
\begin{equation}
\label{eqc3}
\mathbf{g}_{1_{i,j}}^r=-\frac{\ervlfeature(j)}{\sin \Big(\dista \big(\ervlfeature(j), \ercenter\big) \Big) \| \ervlfeature(j) \|_2}
\end{equation}
Then, we can represent the averaged gradient considering the three terms:
\begin{equation}
\label{eqc4}
\Delta \ercenter = \Delta_{\mathbf{g}_{1_{i,j}}^r} + \Delta_{\mathbf{g}_{2_{i,j}}^r} + \Delta_{\mathbf{h}_{i,j}^r}
\end{equation}
For example, $\Delta_{\mathbf{g}_{1_{i,j}}^r}$ is computed as follows:
\begin{equation}
\label{eqc5}
\Delta_{\mathbf{g}_{1_{i,j}}^r} = \frac{1}{N}\Bigg( \frac{\sum_{i:\elabel(j)=1}\mathbf{g}_{1_{i,j}}^r \delta(\scalebox{.9}{$\tilde{\mathcal{L}}_{\text{\atclossname{}}_{i,j}}^r$}>0)}{1+\sum_{i:\elabel(j)=1}\delta(\scalebox{.9}{$\tilde{\mathcal{L}}_{\text{\atclossname{}}_{i,j}}^r$}>0)} \Bigg)
\end{equation}
Here, $\delta(condition)=1$ if the $condition$ is true and $\delta(condition)=0$ otherwise. Finally, the centers are updated using $\Delta \ercenter$ for every iteration of the training process by a gradient descent algorithm. More details can be found in the supplementary material.

\subsection{Adopting an adversarial approach (\ourlossname{})}
We further improve the performance of the proposed \intlossname{} by applying an adversarial approach inspired by ACoL~\cite{zhang2018adversarial}. For each stream, there are two parallel branches that operate in an adversarial way. The motivation is that a network with a single branch might be dominated by salient activity features that are not enough to localize all the activities in time. We zero out the most salient activity features localized by the first branch for activity class $j$ of $\evideo$ as follows:
\begin{equation}
\label{eq9}
\erafeature(t)=
\begin{cases}
\mathbf{0}, & \text{if}\ \ertcam(j,t) \in top \mhyphen k_a \ \text{elements of}\ \ertcam(j)  \\
\erfeature(t), & \text{otherwise}
\end{cases}
\end{equation}
where $\erafeature \in \mathbb{R}^{D \times l_i}$ denotes the input features of activity class $j$ for the second (adversarial) branch and $k_a$ is set to $\floor{\frac{l_i}{s_a}}$ for a hyperparameter $s_a$ that controls the ratio of zeroed-out features. For each activity class $j$, a separate 1-D convolutional layer of the adversarial branch transforms $\erafeature$ to the classification scores of the activity class $j$ over time. By iterating over all the activity classes, new \tcam{} ${\eratcam \in \mathbb{R}^{N_c \times l_i}}$ is computed. We argue that $\eratcam$ can be used to find other supplementary activities that are not localized by the first branch. By using the original features $\erfeature$, new \tcam{} $\eratcam$, and a separate set of centers ${\{\eracenter\}_{j=1}^{N_c}}$, we can compute the loss of \intlossname{} for this adversarial branch $\mathcal{L}_{\text{\intlossname{}}}^{ra}$ in a similar manner (Eq.~\ref{eq1}-\ref{eq7}). We call the sum of the losses of the two branches \ourlossnamelong{} (\ourlossname{}):
\begin{equation}
\label{eq10}
\mathcal{L}_{\text{\ourlossname{}}}^r = \mathcal{L}_{\text{\intlossname{}}}^r + \mathcal{L}_{\text{\intlossname{}}}^{ra}
\end{equation}
In addition, the losses for the optical flow stream $\mathcal{L}_{\text{\intlossname{}}}^o$ and $\mathcal{L}_{\text{\intlossname{}}}^{oa}$ are also computed in the same manner. As a result, the total \ourlossname{} term is given by:
\begin{equation}
\label{eq11}
\mathcal{L}_{\text{\ourlossname{}}} = \mathcal{L}_{\text{\ourlossname{}}}^r+\mathcal{L}_{\text{\ourlossname{}}}^o
\end{equation}

\subsection{Classification Loss} \label{subsec:clsloss}
Following the previous works~\cite{paul2018w,liu2019completeness,narayan20193c}, we use the cross-entropy between the predicted pmf (probability mass function) and the ground-truth pmf of activities for classifying different activity classes in a video. We will first look at the RGB stream. For each video $\evideo$, we compute the class-wise classification scores ${\erscore \in \mathbb{R}^{N_c}}$ by averaging $top \mhyphen k$ elements of $\ertcam$ per activity class where $k$ is set to $\ceil{\frac{l_i}{s}}$ for a hyperparameter $s$. Then, the softmax function is applied to compute the predicted pmf of activities ${\mathbf{p}_i^r \in \mathbb{R}^{N_c}}$. The ground-truth pmf $\mathbf{q}_i$ is obtained by $l_1 \mhyphen \text{normalizing}$ $\elabel$. Then, the classification loss for the RGB stream is:
\begin{equation}
\label{eq12}
\mathcal{L}_{\text{CLS}}^r = \frac{1}{N}\sum_{i=1}^N \sum_{j=1}^{N_c} -\mathbf{q}_i(j)\log\big(\mathbf{p}_i^r(j)\big)
\end{equation}
The classification loss for the optical flow stream $\mathcal{L}_{\text{CLS}}^o$ is computed in a similar manner. $\mathcal{L}_{\text{CLS}}^{ra}$ and $\mathcal{L}_{\text{CLS}}^{oa}$ of adversarial branches are also computed in the same way.

Finally, we compute the final \tcam{} $\eftcam$ from the four \tcam{}s (two from the RGB stream: $\ertcam, \eratcam$, two from the optical flow stream: $\eotcam, \eoatcam$) as follows:
\begin{equation}
\label{eq13}
\eftcam = \wrfortcam \cdot (\ertcam + \omega\eratcam) + \wofortcam \cdot (\eotcam + \omega\eoatcam)
\end{equation}
where $\wrfortcam, \wofortcam \in \mathbb{R}^{N_c}$ are class-specific weighting parameters that are learned during training and $\omega$ is a hyperparameter for the relative importance of \tcam{}s from the adversarial branch. We can then compute the classification loss for the final \tcam{} $\mathcal{L}_{\text{CLS}}^F$ in the same manner. The total classification loss is given by:
\begin{equation}
\label{eq14}
\mathcal{L}_{\text{CLS}} = \mathcal{L}_{\text{CLS}}^r + \mathcal{L}_{\text{CLS}}^{ra} + \mathcal{L}_{\text{CLS}}^o +
\mathcal{L}_{\text{CLS}}^{oa} +
\mathcal{L}_{\text{CLS}}^F
\end{equation}

\subsection{Classification and Localization}
During the test time, we use the final \tcam{} $\eftcam$ for the classification and localization of activities following the previous works~\cite{paul2018w,narayan20193c}. First, we compute the class-wise classification scores ${\efscore \in \mathbb{R}^{N_c}}$ and the predicted pmf of activities ${\mathbf{p}_i^F \in \mathbb{R}^{N_c}}$ as described in Section~\ref{subsec:clsloss}. We use $\mathbf{p}_i^F$ for activity classification. For activity localization, we first find a set of possible activities that has positive classification scores, which is ${\{j: \efscore(j)>0\}}$. For each activity in this set, we localize all the temporal segments that has positive \tcam{} values for two or more successive time steps. Formally, a set of localized temporal segments for $\evideo$ is:
\begin{equation}
\label{eq15}
\{[s,e]: \forall t \in [s,e], \ \eftcam(t)>0 \ \text{and} \ \eftcam(s-1)<0 \ \text{and} \  \eftcam(e+1)<0\}
\end{equation}
where $\eftcam(0)$ and $\eftcam(l_i+1)$ are defined to be any negative values and ${e\geq s+2}$. The localized segments for each activity are non-overlapping. We assign a confidence score for each localized segment, which is the sum of the maximum \tcam{} value of the segment and the classification score of it.

\section{Experiments}
\subsection{Datasets and Evaluation}
We evaluate our method on two datasets: \thumos{}~\cite{THUMOS14} and \actthree{}~\cite{caba2015activitynet}. For the \thumos{} dataset, the validation videos are used for training without temporal boundary annotations and the test videos are used for evaluation following the convention in the literature. This dataset is known to be challenging because each video has a number of activity instances and the duration of the videos varies widely. For the \actthree{} dataset, we use the training set for training and the validation set for evaluation.
Following the standard evaluation protocol, we report mean average precision (mAP) at different intersection over union (IoU) thresholds.

\subsection{Implementation Details}
First, we extract RGB frames from each video at 25 fps and generate optical flow frames by using the TV-L1 algorithm~\cite{zach2007duality}. Each video is then divided into non-overlapping 16-frame segments. We apply I3D networks~\cite{carreira2017quo} pre-trained on Kinetics dataset~\cite{kay2017kinetics} to the segments to obtain the intermediate 1024-dimensional features after the global pooling layer. We train our network in an end-to-end manner using a single GPU (TITAN Xp).

For the \thumos{} dataset~\cite{THUMOS14}, we train our network using a batch size of 32. We use the Adam optimizer~\cite{kingma2014adam} with learning rate $10^{-4}$ and weight decay 0.0005. The centers are updated using the SGD algorithm with learning rate 0.1 for the RGB stream and 0.2 for the optical flow stream. The kernel size of the 1-D convolutional layers for the \tcam{}s is set to 1. We set $\alpha$ in Eq.~\ref{eq1} to 1 and $\gamma$ in Eq.~\ref{eq8} to 0.6. For $\beta$ in Eq.~\ref{eq6}, we randomly generate a number between 0.001 and 0.1 for each training sample. We set angular margins $m_1$ to 2 and $m_2$ to 1. $s_a$ of Eq.~\ref{eq9} and $s$ for the classification loss are set to 40 and 8, respectively. Finally, $\omega$ in Eq.~\ref{eq13} is set to 0.6. The whole training process of 40.5k iterations takes less than 14 hours.

For the \actthree{} dataset~\cite{caba2015activitynet}, it is shown from the previous works~\cite{paul2018w,narayan20193c} that post-processing of the final \tcam{} is required. We use an additional 1-D convolutional layer (kernel size=13, dilation=2) to post-process the final \tcam{}. The kernel size of the 1-D convolutional layers for \tcam{}s is set to 3. In addition, we change the batch size to 24. The learning rate for centers are 0.05 and 0.1 for the RGB and optical flow streams, respectively. We set $\alpha$ to 2, $\gamma$ to 0.2, and $\omega$ to 0.4. The remaining hyperparameters of $\beta$, $m_1$, $m_2$, $s_a$, and $s$ are the same as above. We train the network for 175k iterations.

\subsection{Comparisons with the State-of-the-art}
\begin{table}[t]
\caption{Performance comparison of \ourlossname{} with state-of-the-art methods on the \thumos{} dataset~\cite{THUMOS14}. \ourlossname{} significantly outperforms all the other weakly-supervised methods. $\dagger$ indicates an additional usage of other ground-truth annotations or independently collected data. \ourlossname{} also outperforms all weakly$\dagger$-supervised methods that use additional data at higher IoUs (from 0.4 to 0.9). The column AVG is for the average mAP of IoU threshold from 0.1 to 0.9.}
\label{table1}
\centering
\resizebox{\columnwidth}{!}{%
\begin{tabular}{c|l|cccccccccc}
\hline
\multicolumn{1}{c|}{\multirow{2}{*}{\raisebox{-0.12\height}{\makebox[5.8em]{Supervision}}}} & \multicolumn{1}{c|}{\multirow{2}{*}{\raisebox{-0.12\height}{\makebox[9.1em]{Method}}}} & \multicolumn{10}{c}{\raisebox{-0.12\height}{mAP(\%)@\hspace{0.1em}IoU}} \rule{0pt}{8pt}\\
& & \makebox[2.5em]{0.1} & \makebox[2.5em]{0.2} & \makebox[2.3em]{0.3} & \makebox[2.3em]{0.4} & \makebox[2.3em]{0.5} & \makebox[2.3em]{0.6} & \makebox[2.3em]{0.7} & \makebox[2.3em]{0.8} & \makebox[2.3em]{0.9} & \makebox[3.3em]{AVG}\\
\hline \hline
\rule{0pt}{9.7pt}\parbox[t]{5.6mm}{\multirow{6}{*}{Full}} & \hspace{0.3em}S-CNN~\cite{shou2016temporal} & 47.7 & 43.5 & 36.3 & 28.7 & 19.0 & 10.3 & 5.3 & - & - & - \\
%& \hspace{0.3em}CDC~\cite{shou2017cdc} & -  & -  & 40.1 & 29.4 & 23.3 & 13.1 & 7.9 & - & - & - \\
& \hspace{0.3em}R-C3D~\cite{xu2017r} & 54.5 & 51.5 & 44.8 & 35.6 & 28.9 & - & - & - & - & - \\
& \hspace{0.3em}SSN~\cite{zhao2017temporal} & 66.0 & 59.4 & 51.9 & 41.0 & 29.8 & - & - & - & - & - \\
& \hspace{0.3em}TAL-Net~\cite{chao2018rethinking} & 59.8 & 57.1 & 53.2 & \textbf{48.5} & \textbf{42.8} & \textbf{33.8} & \textbf{20.8} & - & - & - \\
& \hspace{0.3em}BSN~\cite{lin2018bsn} & - & - & 53.5 & 45.0 & 36.9 & 28.4 & 20.0 & - & - & - \\
& \hspace{0.3em}GTAN~\cite{long2019gaussian} & \textbf{69.1} & \textbf{63.7} & \textbf{57.8} & 47.2 & 38.8 & - & - & - & - & - \\
\hline
\rule{0pt}{9.6pt}\parbox[t]{7.6mm}{\multirow{4}{*}{Weak\hspace{0.1em}$\dagger$}} & \hspace{0.3em}Liu \etal~\cite{liu2019completeness} & 57.4 & 50.8 & 41.2 & 32.1 & 23.1 & 15.0 & 7.0 & - & - & - \\
& \hspace{0.3em}3C-Net~\cite{narayan20193c} & 59.1 & 53.5 & 44.2 & 34.1 & 26.6 & - & 8.1 & - & - & - \\
& \hspace{0.3em}Nguyen \etal{}~\cite{nguyen2019weakly} & 64.2 & 59.5 & \textbf{49.1} & \textbf{38.4} & \textbf{27.5} & \textbf{17.3} & \textbf{8.6} & \textbf{3.2} & \textbf{0.5} & \textbf{29.8}\\
& \hspace{0.3em}STAR~\cite{xu2019segregated} & \textbf{68.8} & \textbf{60.0} & 48.7 & 34.7 & 23.0 & - & - & - & - & - \\[0.1pt]
\hline
\rule{0pt}{9.6pt}\parbox[t]{7mm}{\multirow{7}{*}{Weak}} & \hspace{0.3em}UntrimmedNet~\cite{wang2017untrimmednets} & 44.4 & 37.7 & 28.2 & 21.1 & 13.7 & - & - & - & - & - \\
& \hspace{0.3em}STPN~\cite{nguyen2018weakly} & 52.0 & 44.7 & 35.5 & 25.8 & 16.9 & 9.9 & 4.3 & 1.2 & 0.1 & 21.2 \\
& \hspace{0.3em}W-TALC~\cite{paul2018w} & 55.2 & 49.6 & 40.1 & 31.1 & 22.8 & - & 7.6 & - & - & - \\
& \hspace{0.3em}AutoLoc~\cite{shou2018autoloc} & - & - & 35.8 & 29.0 & 21.2 & 13.4 & 5.8 & - & - & - \\
& \hspace{0.3em}CleanNet~\cite{liu2019weakly} & - & - & 37.0 & 30.9 & 23.9 & 13.9 & 7.1 & - & - & - \\
& \hspace{0.3em}MAAN~\cite{yuan2018marginalized} & 59.8 & 50.8 & 41.1 & 30.6 & 20.3 & 12.0 & 6.9 & 2.6 & 0.2 & 24.9 \\
& \hspace{0.3em}BaS-Net~\cite{lee2020background} & 58.2 & 52.3 & 44.6 & 36.0 & 27.0 & 18.6 & 10.4 & 3.9 & 0.5 & 27.9 \\ \cline{2-12}
\rule{0pt}{9.6pt} & \hspace{0.3em}\ourlossname{} (Ours) & \textbf{61.2} & \textbf{56.1} & \textbf{48.1} & \textbf{39.0} & \textbf{30.1} & \textbf{19.2} & \textbf{10.6} & \textbf{4.8} & \textbf{1.0} & \textbf{30.0} \\
\hline
\end{tabular}
}
\end{table}

We compare our final method \ourlossname{} with other state-of-the-art approaches on the \thumos{} dataset~\cite{THUMOS14} in Table~\ref{table1}. Full supervision refers to training from frame-level activity annotations, whereas weak supervision indicates training only from video-level activity labels. For fair comparison, we use the symbol $\dagger$ to separate methods utilizing additional ground-truth annotations~\cite{narayan20193c,xu2019segregated} or independently collected data~\cite{liu2019completeness,nguyen2019weakly}. The column AVG is for the average mAP of IoU thresholds from 0.1 to 0.9 with a step size of 0.1. Our method significantly outperforms other weakly-supervised methods across all metrics. Specifically, an absolute gain of 2.1\% is achieved in terms of the average mAP when compared to the best previous method (BaS-Net~\cite{lee2020background}). We want to note that our method performs even better than the methods of weak$\dagger$ supervision at higher IoUs.

We also evaluate \ourlossname{} on the \actthree{} dataset~\cite{caba2015activitynet}. Following the standard evaluation protocol of the dataset, we report mAP at different IoU thresholds, which are from 0.05 to 0.95. As shown in Table~\ref{table2}, our method again achieves the best performance.

\begin{table}[t]
\caption{Performance comparison on the \actthree{} dataset~\cite{caba2015activitynet}. \ourlossname{} again achieves the best performance. $\dagger$ indicates an additional usage of other ground-truth annotations or independently collected data. The column AVG is for the average mAP of IoU threshold from 0.5 to 0.95.}
\label{table2}
\centering
\resizebox{\columnwidth}{!}{%
\begin{tabular}{c|l|ccccccccccc}
\hline
\multicolumn{1}{c|}{\multirow{2}{*}{\raisebox{-0.12\height}{\makebox[5.8em]{Supervision}}}} & \multicolumn{1}{c|}{\multirow{2}{*}{\raisebox{-0.12\height}{\makebox[8.5em]{Method}}}} & \multicolumn{11}{c}{\raisebox{-0.12\height}{mAP(\%)@\hspace{0.1em}IoU}} \rule{0pt}{8pt}\\
& & \makebox[2.5em]{0.5} & \makebox[2.5em]{0.55} & \makebox[2.3em]{0.6} & \makebox[2.3em]{0.65} & \makebox[2.3em]{0.7} & \makebox[2.3em]{0.75} & \makebox[2.3em]{0.8} & \makebox[2.3em]{0.85} & \makebox[2.3em]{0.9} & \makebox[2.3em]{0.95} & \makebox[3.3em]{AVG}\\
\hline \hline
\rule{0pt}{9.7pt}\parbox[t]{7.6mm}{\multirow{3}{*}{Weak\hspace{0.1em}$\dagger$}} & \hspace{0.3em}Liu \etal~\cite{liu2019completeness} & 34.0 & - & - & - & - & \textbf{20.9} & - & - & - & \textbf{5.7} & \textbf{21.2} \\
& \hspace{0.3em}Nguyen \etal{}~\cite{nguyen2019weakly} & \textbf{36.4} & - & - & - & - & 19.2 & - & - & - & 2.9 & - \\
& \hspace{0.3em}STAR~\cite{xu2019segregated} & 31.1 & - & - & - & - & 18.8 & - & - & - & 4.7 & - \\[0.1pt]
\hline
\rule{0pt}{9.6pt}\parbox[t]{7mm}{\multirow{4}{*}{Weak}} & \hspace{0.3em}STPN~\cite{nguyen2018weakly} & 29.3 & - & - & - & - & 16.9 & - & - & - & 2.6 & - \\
& \hspace{0.3em}MAAN~\cite{yuan2018marginalized} & 33.7 & - & - & - & - & 21.9 & - & - & - & \textbf{5.5} & - \\
& \hspace{0.3em}BaS-Net~\cite{lee2020background} & 34.5 & - & - & - & - & \textbf{22.5} & - & - & - & 4.9 & 22.2 \\ \cline{2-13}
\rule{0pt}{9.6pt} & \hspace{0.3em}\ourlossname{} (Ours) & \textbf{36.8} & 33.6 & 30.8 & 27.8 & 24.9 & 22.0 & 18.1 & 14.9 & 10.2 & 5.2 & \textbf{22.5} \\
\hline
\end{tabular}
}
\end{table}

\subsection{Ablation Study and Analysis}
We perform an ablation study on the \thumos{} dataset~\cite{THUMOS14}. In Table~\ref{table3}, we analyze the two main contributions of this work, which are the usage of the newly-suggested triplet (Eq.~\ref{eq7}) and the adoption of adversarial approach (Eq.~\ref{eq10}). \atclossname{} refers to the baseline that uses only the loss term of Eq.~\ref{eq4}. We use the superscript + to indicate the addition of adversarial branch. As described in Section~\ref{subsec:adot}, \intlossname{} additionally uses the new triplet on top of the baseline. We can observe that our final proposed method, \ourlossname{}, performs the best. It implies that both components are necessary to achieve the best performance and each of them is effective. Interestingly, adding an adversarial branch does not bring any performance gain without our new triplet. We think that although using \intlossname{} increases the localization performance by learning discriminative features, it also makes the network sensitive to salient activity-related features.

\begin{figure}[t]
  \adjustbox{valign=t}{\begin{minipage}[t]{0.47\columnwidth}
  \small
    \includegraphics[width=0.98\columnwidth]{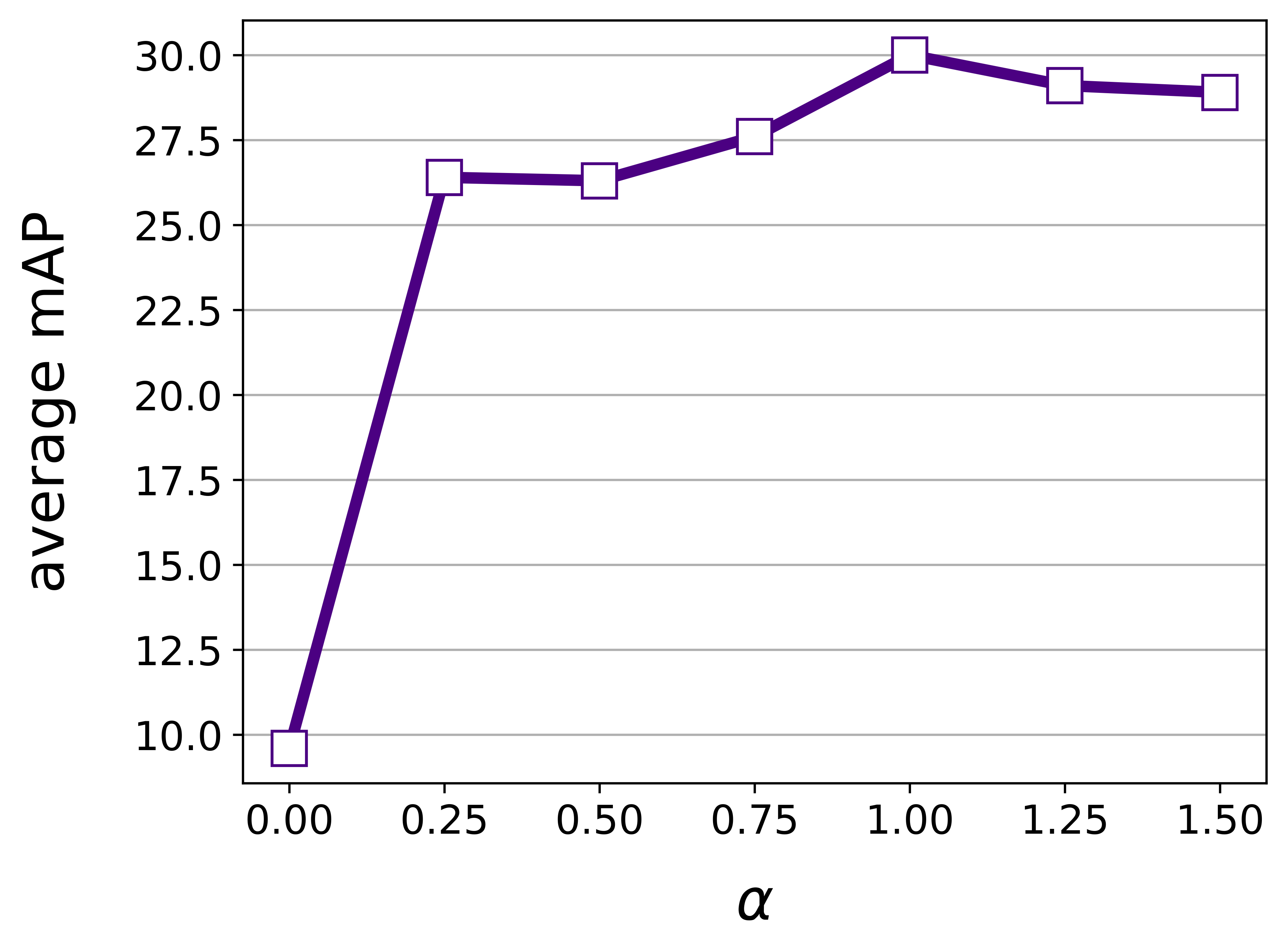}\\[-0.5ex] \hspace*{9.5em}(a)
  \end{minipage}}
  \hspace{0.02cm}
  \adjustbox{valign=t}{\begin{minipage}[t]{0.47\columnwidth}
  \small
    \includegraphics[width=0.98\columnwidth]{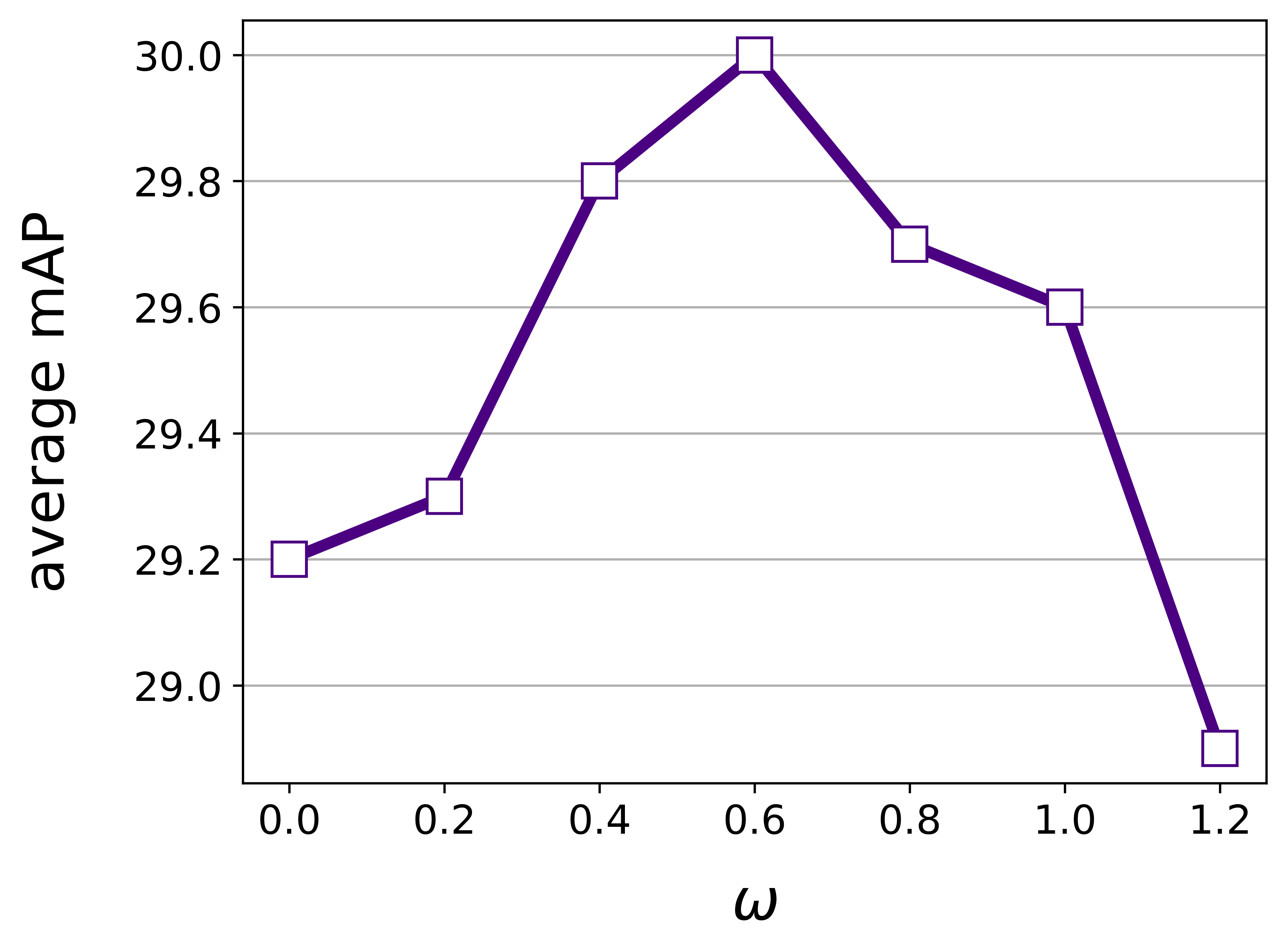}\\[-0.5ex] \hspace*{9.48em}(b)
  \end{minipage}}
  \caption{We analyze the impact of two main hyperparameters $\alpha$ and $\omega$. (a): Positive $\alpha$ always provides the performance gain, so it indicates that our method is effective. (b): If $\omega$ is too large, the performance is decreased substantially. It implies that \tcam{}s of adversarial branches provide mostly supplementary information.}
  \label{fig3}
\end{figure}

We analyze the impact of two main hyperparameters in Fig.~\ref{fig3}. The first one is $\alpha$ that controls the weight of \ourlossname{} term (Eq.~\ref{eq1}), and the other one is $\omega$ that is for the relative importance of \tcam{}s from adversarial branches (Eq.~\ref{eq13}). We can observe from Fig.~\ref{fig3}(a) that positive $\alpha$ always brings the performance gain. It indicates that \ourlossname{} is effective. As seen in Fig.~\ref{fig3}(b), the performance is increased by using an adversarial approach when $\omega$ is less or equal to 1. If $\omega$ is greater than 1, \tcam{}s of adversarial branches will play a dominant role in activity localization. Therefore, the results tell us that the adversarial branches provide mostly supplementary information.

\begin{table}[t]
\caption{Performance comparison of different ablative settings on the \thumos{} dataset~\cite{THUMOS14}. The superscript + indicates that we add an adversarial branch to the baseline method. It demonstrates that both components are effective.}
\label{table3}
\centering
\resizebox{0.95\columnwidth}{!}{%
\begin{tabular}{l|c|c||cccccc}
\hline
\multicolumn{1}{c|}{\multirow{2}{*}{\raisebox{-0.12\height}{\makebox[5.8em]{Method}}}} &
\multicolumn{1}{c|}{\multirow{2}{*}{\raisebox{-0.12\height}{\makebox[6.1em]{New triplet}}}} &
\multicolumn{1}{c||}{\multirow{2}{*}{\raisebox{-0.12\height}{\makebox[6.1em]{\hspace{-0.1em}Adversarial}}}} &
\multicolumn{6}{c}{\raisebox{-0.12\height}{\hspace{0.2em}mAP(\%)@\hspace{0.1em}IoU}} \rule{0pt}{8pt}\\
& & & \hspace{0.2em}\makebox[2.3em]{0.3} & \makebox[2.3em]{0.4} & \makebox[2.3em]{0.5} & \makebox[2.3em]{0.6} & \makebox[2.3em]{0.7} & \makebox[5.6em]{AVG(0.1:0.9)} \\
\hline
\rule{0pt}{9pt}\hspace{0.65em}\atclossname{} & & & \hspace{0.2em}44.7 & 34.8 & 25.7 & 15.8 & 8.3 & 27.4 \\
\hspace{0.65em}$\text{\atclossname{}}^+$ & & $\checkmark$ & \hspace{0.2em}43.7 & 35.1 & 26.3 & 15.7 & 8.3 & 27.2 \\
\hspace{0.65em}\intlossname{} & $\checkmark$ & & \hspace{0.2em}46.6 & 37.2 & 28.9 & 18.2 & 10.0 & 29.2 \\
\hspace{0.65em}\ourlossname{} & $\checkmark$ & $\checkmark$ & \hspace{0.2em}\textbf{48.1} & \textbf{39.0} & \textbf{30.1} & \textbf{19.2} & \textbf{10.6} & \textbf{30.0} \\
\hline
\end{tabular}
}
\end{table}

\subsection{Qualitative Analysis}
\begin{figure}[htbp]
  \small
    \includegraphics[width=\columnwidth]{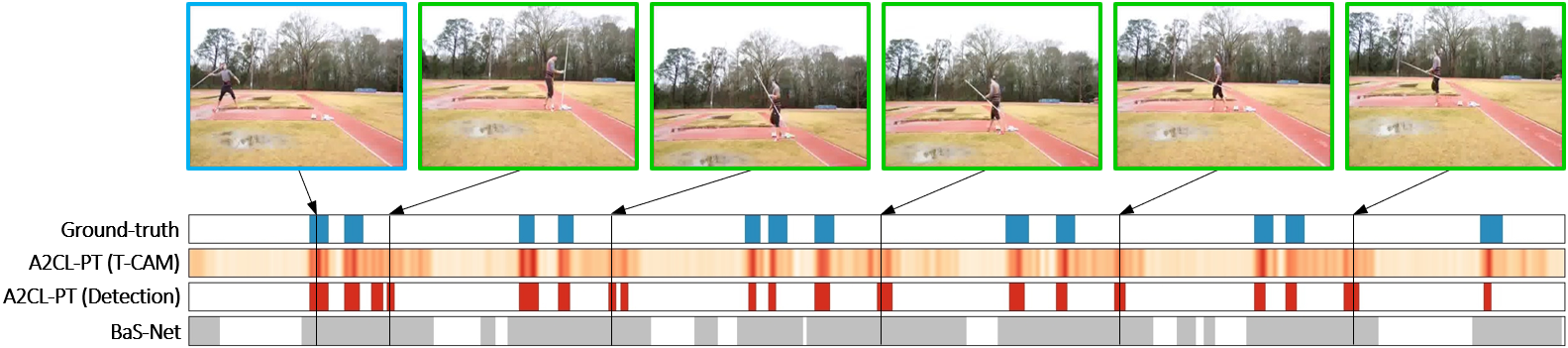}\\[-0.8ex] \hspace*{20.4em}(a) \\[1ex]
    \includegraphics[width=\columnwidth]{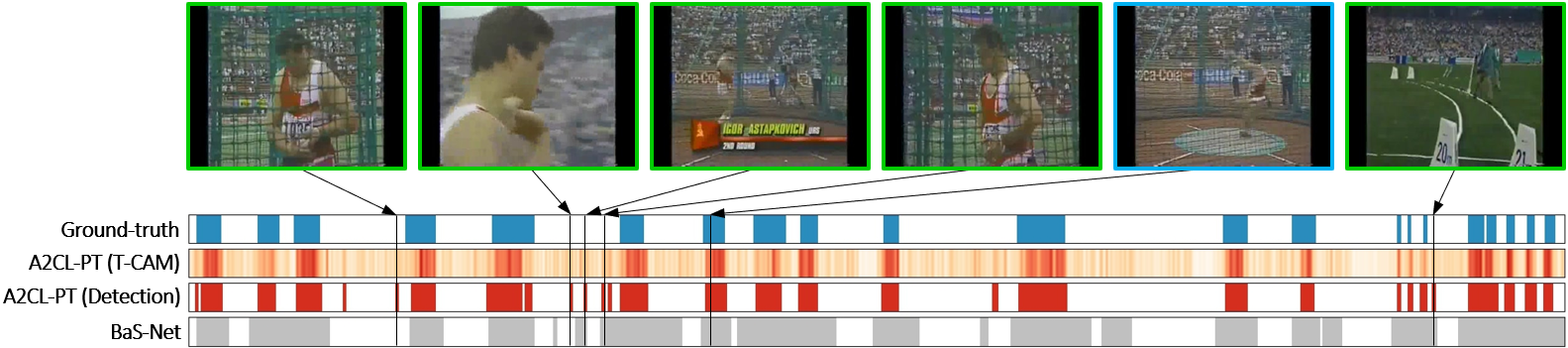}\\[-0.8ex] \hspace*{20.4em}(b) \\[1ex]
    \includegraphics[width=\columnwidth]{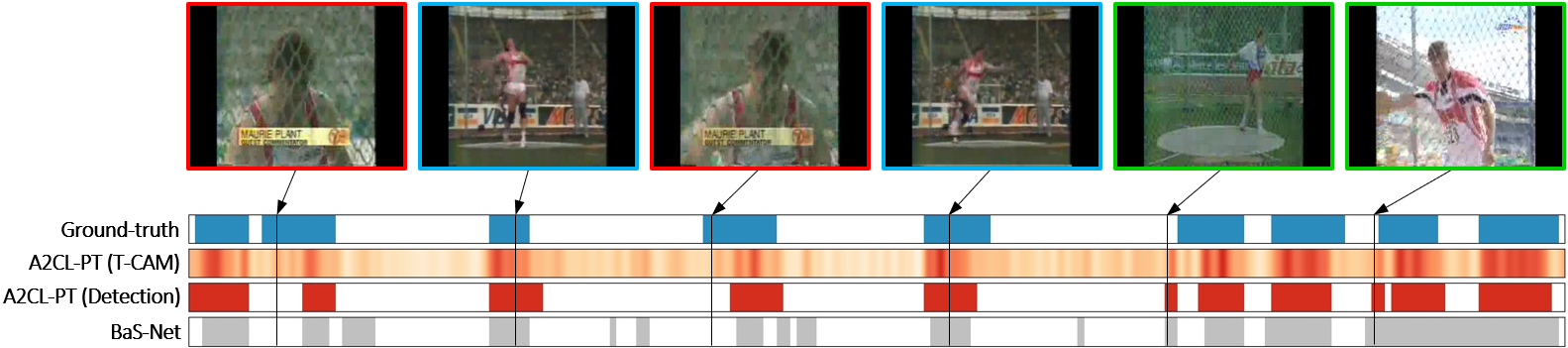}\\[-0.8ex] \hspace*{20.4em}(c) \\[1ex]
    \includegraphics[width=\columnwidth]{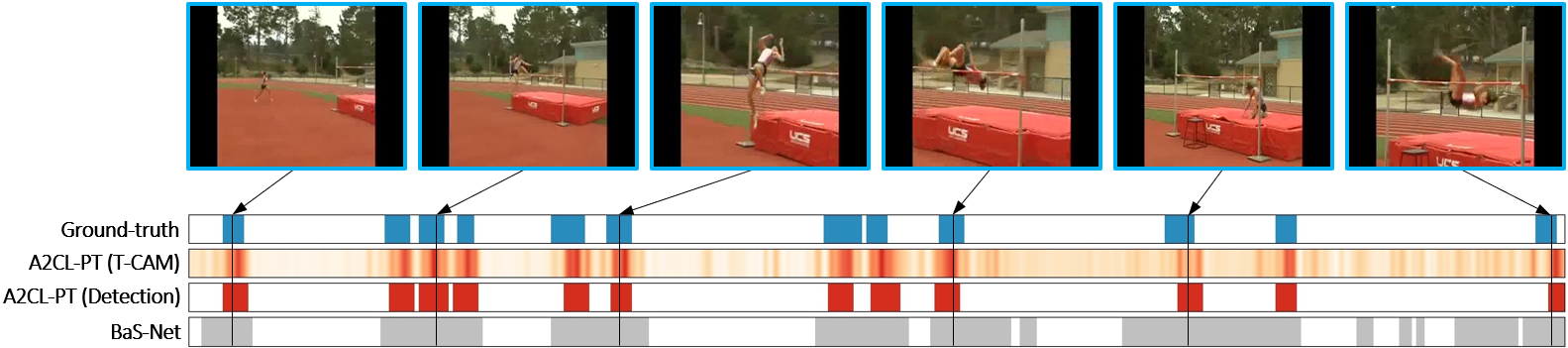}\\[-0.8ex] \hspace*{20.4em}(d)
  \caption{Qualitative results on the \thumos{} dataset~\cite{THUMOS14}. Detection denotes the localized activity segments. The results of \mbox{BaS-Net}~\cite{lee2020background} are also included for additional comparison. Contours of the sampled frames have three different colors. We use blue, green, and red to indicate true positives, false positives, and false negatives, respectively. (a): An example of \textit{JavelinThrow} activity class. The observed false positives are challenging. The person in the video swings the javelin on the frames of these false positives, which can be mistaken for a throw. (b): An example of \textit{HammerThrow}. One of the false positives include the person who draws the line on the field. It is hard to distinguish the two activities. (c): An example of \textit{ThrowDiscus}. Multiple false negatives are observed, which illustrates the situations where the ground-truth activity instances are wrongly annotated. (d): An example of \textit{HighJump} without such unusual cases. It can be observed that our method performs well in general.}
  \label{fig4}
\end{figure}

We perform a qualitative analysis to better understand our method. In Fig.~\ref{fig4}, qualitative results of our \ourlossname{} on four videos from the test set of the \thumos{} dataset~\cite{THUMOS14} are presented. (a), (b), (c), and (d) are examples of \textit{JavelinThrow}, \textit{HammerThrow}, \textit{ThrowDiscus}, and \textit{HighJump}, respectively. Detection denotes the localized activity segments. For additional comparison, we also show the results of \mbox{BaS-Net}~\cite{lee2020background}, which is the leading state-of-the-art method. We use three different colors on the contours of sampled frames: blue, green, and red which denote true positive, false positive, and false negative, respectively. In (a), there are multiple instances of false positive. These false positives are challenging because the person in the video swings the javelin, which can be mistaken for a throw. Similar cases are observed in (b). One of the false positives includes the person drawing the line on the field, which looks similar to a \textit{HammerThrow} activity. In (c), some false negative segments are observed. Interestingly, this is because the ground-truth annotations are wrong; that is, the \textit{ThrowDiscus} activity is annotated but it does not actually occur in these cases. In (d), all the instances of the \textit{HighJump} activity are successfully localized. Other than the unusual situations, our method performs well in general.

\section{Conclusion}
We have presented \ourlossname{} as a novel method for weakly-supervised temporal activity localization. We suggest using two triplets of vectors of the feature space to learn discriminative features and to distinguish background portions from activity-related parts of a video. We also propose to adopt an adversarial approach to localize activities more thoroughly. We perform extensive experiments to show that our method is effective. \ourlossname{} outperforms all the existing state-of-the-art methods on major datasets. Ablation study demonstrates that both contributions are significant. Finally, we qualitatively analyze the effectiveness of our method in detail.

\bigskip

\noindent \textbf{Acknowledgement } We thank Stephan Lemmer, Victoria Florence, Nathan Louis, and Christina Jung for their valuable feedback and comments. This research was, in part, supported by NIST grant 60NANB17D191.

% ---- Bibliography ----
%
% BibTeX users should specify bibliography style 'splncs04'.
% References will then be sorted and formatted in the correct style.
%
\bibliographystyle{splncs04}
\bibliography{ref}
\end{document}